\documentclass[a4paper, 10pt, conference]{ieeeconf}      

\IEEEoverridecommandlockouts                              

\usepackage{amsmath} 
\usepackage{amssymb}  
\usepackage{subcaption}
\usepackage{array}
\usepackage{multirow, makecell}
\usepackage{gensymb}
\usepackage{siunitx}    
\usepackage{tikz}
\usetikzlibrary{fit, shapes, positioning, patterns}

\DeclareMathOperator{\rmse}{RMSE}
\DeclareMathOperator{\tr}{Tr}
\DeclareMathOperator{\sr}{SR}

\newcommand\copyrighttext{%
	\footnotesize \copyright{ }2020 IEEE. Personal use of this material is permitted. Permission from IEEE must be obtained for all other uses, in any current or future media, including reprinting/republishing this material for advertising or promotional purposes, creating new collective works, for resale or redistribution to servers or lists, or reuse of any copyrighted component of this work in other works.}
\newcommand\copyrightnotice{%
	\begin{tikzpicture}[remember picture,overlay]
	\node[anchor=south,yshift=10pt,xshift=7pt] at (current page.south) {\parbox{\dimexpr\textwidth-\fboxsep-\fboxrule\relax}{\copyrighttext}};
	\end{tikzpicture}%
}

\title{\LARGE \bf
Real-Time Point Cloud Fusion of Multi-LiDAR Infrastructure Sensor Setups with Unknown Spatial Location and Orientation*
}

\author{Laurent Kloeker$^{1}$, Christian Kotulla$^{1}$ and Lutz Eckstein$^{1}$
\thanks{*The research leading to these results is funded by the European Regional Development Fund (ERDF) within the project “HDV-Mess - High-precision digital traffic recording as a basis for future mobility research - Construction of mobile and modular measuring stations”. The authors would like to thank the consortium for the successful cooperation.}
\thanks{$^{1}$The authors are with the research area Vehicle Intelligence \& Automated Driving, Institute for Automotive Engineering, RWTH Aachen University, 52074 Aachen, Germany
        {\tt\small laurent.kloeker@ika.rwth-aachen.de, christian.niklas.kotulla@rwth-aachen.de, lutz.eckstein@ika.rwth-aachen.de}}%
}

\begin{document}

\maketitle
\thispagestyle{empty}
\pagestyle{empty}
\copyrightnotice

\begin{abstract}

The use of infrastructure sensor technology for traffic detection has already been proven several times. However, extrinsic sensor calibration is still a challenge for the operator. While previous approaches are unable to calibrate the sensors without the use of reference objects in the sensor field of view (FOV), we present an algorithm that is completely detached from external assistance and runs fully automatically. Our method focuses on the high-precision fusion of LiDAR point clouds and is evaluated in simulation as well as on real measurements. We set the LiDARs in a continuous pendulum motion in order to simulate real-world operation as closely as possible and to increase the demands on the algorithm. However, it does not receive any information about the initial spatial location and orientation of the LiDARs throughout the entire measurement period. Experiments in simulation as well as with real measurements have shown that our algorithm performs a continuous point cloud registration of up to four 64-layer LiDARs in real-time. The averaged resulting translational error is within a few centimeters and the averaged error in rotation is below \num[detect-weight=true]{0.15}~degrees.

\end{abstract}

\section{INTRODUCTION}

The use of infrastructure sensors to record road users is a common method. In addition to pure traffic surveillance, e.g. counting of road users, infrastructure sensors can also be used for the collection of high-precision driving data \cite{Semertzidis2010, Meissner2010}. In both cases, extrinsic calibration of the sensors used is indispensable \cite{Datondji2016}. Focusing on the use of LiDAR sensors, the need for a calibration target within the sensor field of view (FOV) becomes apparent in previous research approaches. For example, three-dimensional, pre-defined objects are used in \cite{Meissner2010, Kummerle2018}, while the algorithm in \cite{Muller2019} utilizes a specially equipped research vehicle that transmits its position via V2X messages.

\begin{figure}
    \centering
    \begin{subfigure}{0.48\columnwidth}
        \includegraphics[width=\columnwidth]{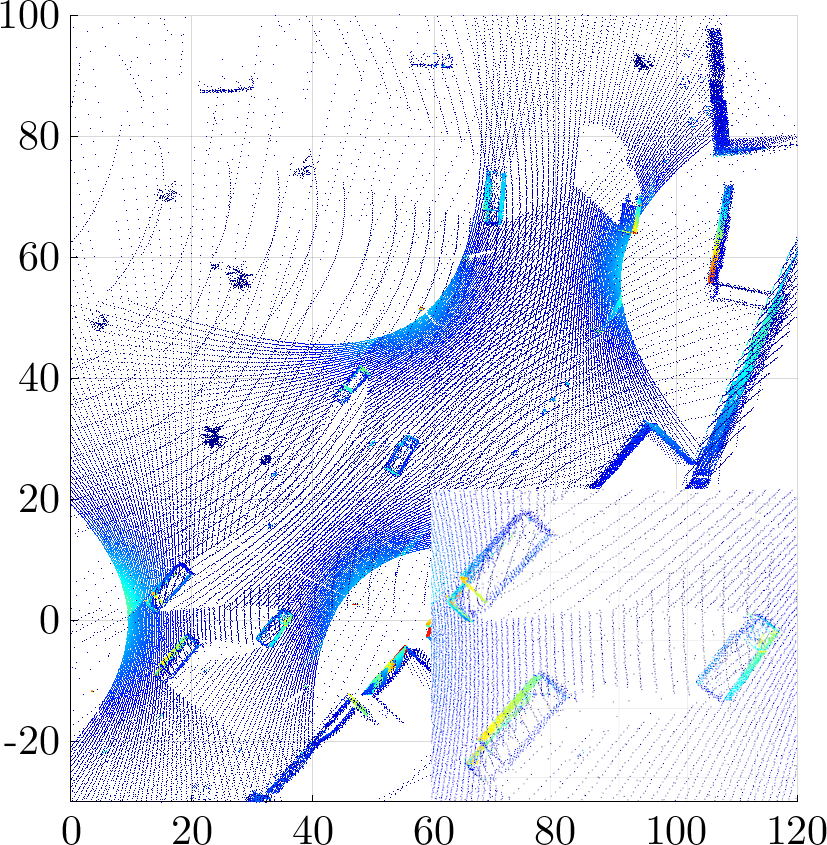}
        \caption{program output (real-time)}
    \end{subfigure}
    \begin{subfigure}{0.48\columnwidth}
        \includegraphics[width=\columnwidth]{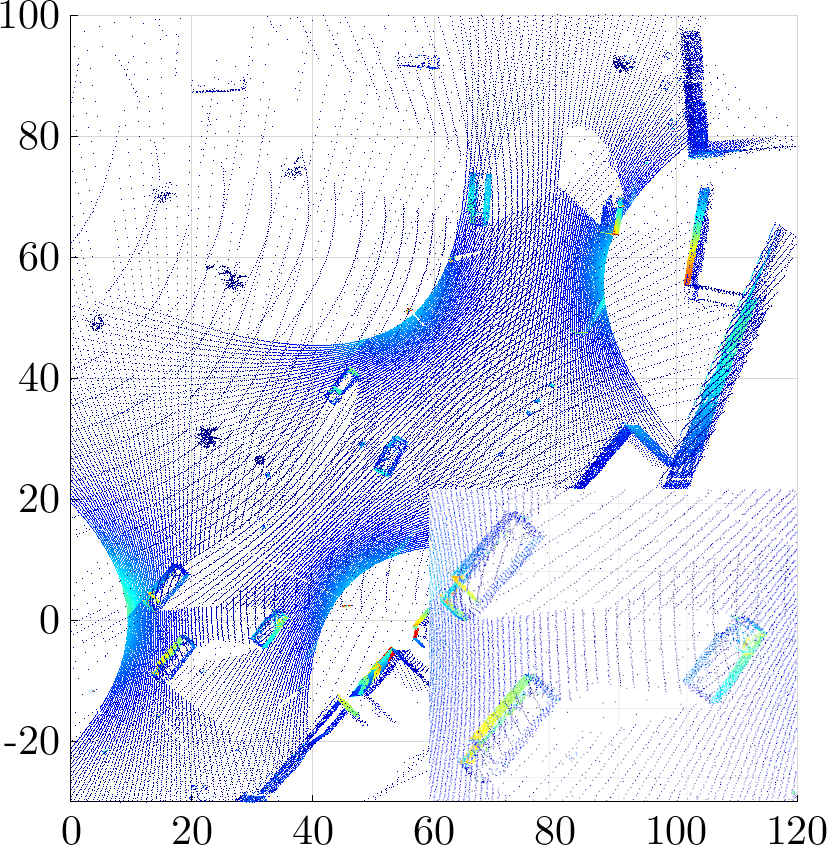}
        \caption{ground truth}
    \end{subfigure}
    \caption{Qualitative comparison between program output and ground truth including close-up views of trucks in the bottom part for one of our simulated scenarios.}
    \label{fig:comparison_fusion_result_and_gt}
\end{figure}
Infrastructure sensors can be designed in the form of mobile or stationary measuring stations. The research project \emph{HDV-Mess}, in whose context this work is being carried out, focuses on the installation of mobile measuring stations for high-precision digital recording of road users \cite{Hdvmess2020}. For both mobile and stationary measuring stations, it is essential that the sensor modules are installed in an elevated position of several meters to minimize object occlusion. However, when sensor modules are installed at such heights, no static positioning of the sensor system can be guaranteed over the measurement period because of external influences such as wind and passing vehicles. An additional improvement is the installation of several measuring stations per road segment in order to record the measuring cross section from several perspectives. Subsequent fusion of all recorded raw sensor data generates a three-dimensional image of the measurement cross section, which considerably reduces the probability of occluded road users. The fusion of the raw sensor data provides considerably more information for each road user per unit of volume or unit area of the measured cross section. The use of deep learning algorithms on such fused raw sensor data for the detection and classification of road users is much more promising than their use on the limited raw data of individual sensors due to the significantly higher information density. However, the fusion of raw sensor data requires that the exact position and orientation of each sensor in the entire measuring cross section is known exactly at each time step of the recording. Such a permanent exact determination is not feasible in real operation if performed manually and represents a challenge. Sensor modules for recording traffic data usually consist of cameras, RADAR sensors and LiDAR sensors. Within the scope of this work, we focus on the use of LiDARs as infrastructure sensors. If a LiDAR moves a few centimeters during a measurement, this can lead to deviations of several meters in the measured point cloud, depending on the sensor's range of vision. Such enormous deviations make it impossible to provide reliable information about the positions of the road users recorded. In addition, environmental influences such as wind can cause the sensor modules to oscillate and shift the sensors in translation and rotation by an unknown value at each time step. Besides the unknown position and orientation of the sensor modules during the measurements, the exact initial orientation of all sensor modules to each other is also unknown. Although the positions of the installed sensors can be roughly estimated with considerable effort, measurement errors in the two-digit centimeter range  cannot be excluded. This results in considerable inaccuracy of the recorded measurement data. A highly accurate fusion of the LiDAR point clouds of all installed sensor modules per measurement cross section is therefore indispensable when highly accurate traffic data is required.

Within this work we present an algorithm that determines the initial as well as the continuous unknown spatial position and orientation of several LiDAR point clouds. Subsequently, the algorithm fuses the point clouds. Our algorithm is tested on simulated data and real-world data. The simulated data consist of generated LiDAR point clouds in digital urban environments, which are augmented with sensor inaccuracies as well as motion inaccuracies. The real-world data were recorded using \emph{Ouster~OS1} LiDARs with \num{64}~layers in a real test environment. The sampling rate for all scenarios is \num{20}~hertz. In both cases, our fusion algorithm is applied to all recorded LiDAR point clouds. We determine the fusion accuracy by comparing the calculated fusion matrices with the ground truth (GT). The determination of the GT in the simulation data is the output of the actual sensor positions and orientations at each measurement time step. For the determination of the GT in the real data we present another approach, which is based on known sensor positions in a geo-referenced orthophoto. After a qualitative and quantitative evaluation of the fusion results, we finally evaluate the required runtimes to prove the real-time capability of our algorithm. Our main contributions are as follows:

\begin{itemize}
    \item We are building an algorithm that can initially and continuously fuse up to four 64-layer LiDARs of unknown spatial location and orientation installed on a measurement cross section with high accuracy in real-time.
    \item We are evaluating our fusion algorithm and prove its real-time capability both in complex simulations and under field conditions with real sensors.
\end{itemize}

\section{RELATED WORK}

Monitoring traffic at road segments is an active research topic. In \cite{Tarko2017} a stationary vehicle with a telescoping mast is used to detect and track vehicles on the road. However, the method requires considerable preparatory work and only uses a single LiDAR. This makes the system prone to occlusions in the scene and limits its monitoring abilities. To overcome this limitation, a setup consisting of multiple sensors can be used. For example in \cite{Kummerle2018}, multiple LiDARs as well as cameras are calibrated based on the trajectory of a spherical registration target. This algorithm achieves good registration results, but depends on the registration target being moved and recognized within each sensor's FOV. For infrastructure sensors, this overlapping area often corresponds to the street being monitored, so a temporary road closure might be necessary. A different method that uses cooperative awareness messages (CAM) for calibration is shown in \cite{Muller2019}. The CAM messages are sent by a specific vehicle (SV) that is used as the registration target. While this algorithm achieves promising results it is also dependent on a calibration target, namely the SV. Further it relies on the assumption that the scene can be monitored without any road user (except for the SV) for a short time interval. Finding appropriate time slots to prevent a road closure can be extremely difficult for busy streets or intersections with multiple lanes for every direction or pedestrians in the scene. Our algorithm does not need any calibration object and can directly be applied to dynamic scenes.

\section{METHOD}

In this section the registration method is described. After formulating the problem, the registration procedure is explained. This is divided into two parts, the initial and continuous registration.

\subsection{Problem definition}

The observed area is monitored from different perspectives using $n_L$ different LiDAR sensors $L_i$ with $i = 1, ..., n_L$. The resulting point clouds $P_i$ are stored in the corresponding local sensor coordinate systems that differ between sensors. We call the set of point clouds generated by the sensors at a specific time step a frame. For further processing, e.g. object detection and tracking, all measurements have to be transformed into the same common coordinate system for every frame. Thus the pose, i.e. position and orientation, of each sensor w.r.t. this common coordinate system must be known. This corresponds to finding relative rigid transformations
\begin{equation}
T_{i,f}^{j,g} =
\begin{bmatrix}
R_{i,f}^{j,g} & t_{i,f}^{j,g} \\
0 & 1
\end{bmatrix}
\in SE(3)
\end{equation}
from the source sensor coordinate system of LiDAR $i$ of frame $f$ to the target sensor $j$ of frame $g$. These consist of a rotation matrix $R_{i,f}^{j,g} \in SO(3)$ and a translation $t_{i,f}^{j,g} \nobreak\in\nobreak \mathbb{R}^3$, resulting in six degrees of freedom (DOF). $SE(3)$ and $SO(3)$ denote the special Euclidean and special orthogonal group respectively. Without loss of generality $L_0$ is chosen as the root sensor and registration target.

\subsection{Initial Registration}

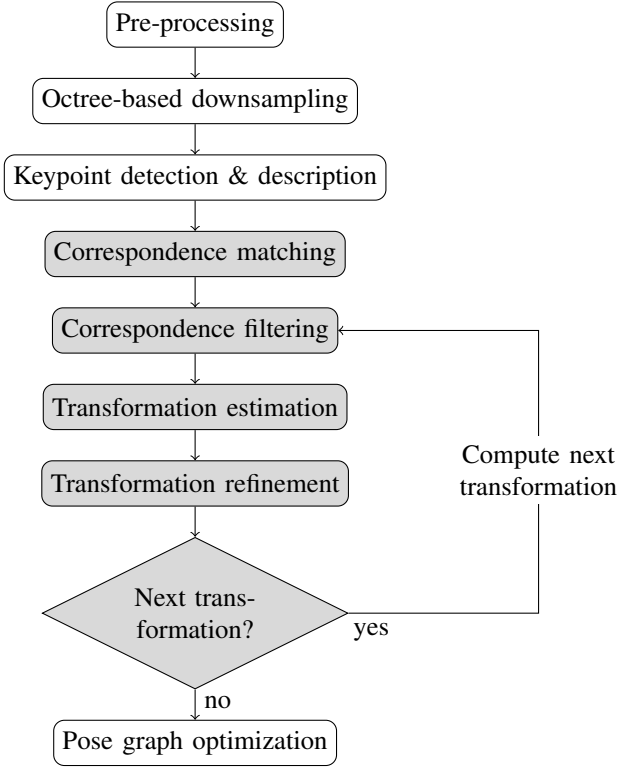
\begin{figure}
    \centering
    \begin{tikzpicture}
        \tikzstyle{flow} = [draw, rectangle, rounded corners, minimum height=0.6cm]
        \tikzstyle{hatched} = [fill=gray!30] 
        
        \node [flow] (preprocessing) {Pre-processing};
        \node [flow] (downsampling) [below=0.4cm of preprocessing] {Octree-based downsampling};
        \node [flow, below=0.4cm of downsampling] (sift_fpfh) {Keypoint detection \& description};
        \node [flow, below=0.4cm of sift_fpfh, hatched] (correspondence_matching) {Correspondence matching};
        \node [flow, below=0.4cm of correspondence_matching, hatched] (correspondence_filtering) {Correspondence filtering};
        \node [flow, below=0.4cm of correspondence_filtering, hatched] (transformation_estimation) {Transformation estimation};
        \node [flow, below=0.4cm of transformation_estimation, hatched] (transformation_refinement) {Transformation refinement};
        \node [draw,diamond,aspect=2,text width=2cm,align=center,below=0.4cm of transformation_refinement, hatched] (next_transformation) {Next transformation?};
        \node[flow,below=0.4cm of next_transformation] (pose_graph_optimization) {Pose graph optimization};
        
        \draw[->] (preprocessing) -- (downsampling);
        \draw[->] (downsampling) -- (sift_fpfh);
        \draw[->] (sift_fpfh) -- (correspondence_matching);
        \draw[->] (correspondence_matching) -- (correspondence_filtering);
        \draw[->] (correspondence_filtering) -- (transformation_estimation);
        \draw[->] (transformation_estimation) -- (transformation_refinement);
        \draw[->] (transformation_refinement) -- (next_transformation);
        
        \draw[->] (next_transformation.east) -|++ (2.5cm, 0cm) node [pos=0.06,below] {yes} --++ (0cm, 1.9cm) node [fill=white, align=center] {Compute next\\ transformation} |- (correspondence_filtering);
        \draw[->] (next_transformation.south) -- node [pos=0.4,right] {no} (pose_graph_optimization);
    \end{tikzpicture}
    \caption{Registration pipeline of initial registration applied to the first frame. Steps with a gray filling are performed for every pair of point clouds within the frame.}
    \label{fig:initial_algorithm_overview}
\end{figure}

The initial registration is used to determine the sensor poses from the data of the first frame using a feature-based approach. First, all sensor streams are synchronized based on their timestamps. After synchronization the maximal temporal offset is $\sr^{-1}$ seconds, where SR is the sample rate. The most important steps of the initial registration pipeline are shown in Fig. \ref{fig:initial_algorithm_overview}.

After pre-processing the point cloud data and removing outliers it is downsampled using an extended variant of the octree-based filtering method in \cite{Zandara2013}. We keep the idea of filtering less relevant areas by creating an octree holding the point cloud and by representing the set of points in every voxel by its centroid. However, the original voxel splitting criterion used (thresholding by z-coordinate) is not sufficient for the given case with unknown sensor orientation. Instead, the normal vectors in every voxel are analyzed. Many points with a similar normal vector orientation indicate planar regions that are less useful for feature-based registration. Thus the octree resolution is only increased by splitting a voxel if the ratio of normals similar to the dominant orientation is smaller than a threshold. To obtain the dominant direction, each normal vector votes for a cell and its surrounding cells on a discretized unit sphere. This keeps feature-rich regions and reduces the differences in point density in feature-less regions. The maximum voxel size is limited to ensure a minimum remaining point density.

The feature-based registration computes all pairwise transformations between different sensors $L_i$ and $L_j$ with $i > j$ at time step 0. The remaining transformations are given by symmetry as $T_{j,0}^{i,0} = (T_{i,0}^{j,0})^{-1}$.
%
%
We compute SIFT keypoints, because their scale-invariance collaborates well with the measuring principle of LiDARs, while being repeatable and invariant to translation and rotation. This has shown to be beneficial in our experiments. The SIFT keypoint detector operates on the surface curvature information from the point normals. The resulting keypoints are then processed by the fast point feature histogram (FPFH) feature descriptor \cite{Rusu2009, Rusu2009a}. Both steps are applied on the downsampled point clouds. The correspondences are obtained by matching the FPFH feature vectors. For a robust transformation estimation a RANSAC-approach is applied to find reliable correspondences. Using a singular value decomposition (SVD), the transformation $T_{i,0}^{j,0}$ can then be estimated from the resulting set of filtered correspondences. A simple check is added to prevent transformations that align the ground planes of two point clouds by flipping the source cloud upside down. This can be avoided by calculating the transformed up-vector
\begin{equation}
    \vec{u}_T = T_{i,0}^{j,0}\,[0,0,1,0]^{\top}
\end{equation}
and discarding $T_{i,0}^{j,0}$ if the z-coordinate of $\vec{u}_T$ is negative. That introduces the constraint that the sensors must not be mounted upside-down, which is no restriction in practical use cases.

After the coarse feature-based registration the result is refined by applying the Generalized Iterative Closest Point (G-ICP) algorithm \cite{Segal2009} to the relative transformations.

Challenges in the datasets like strong viewpoint changes, occlusion, symmetry and partially overlapping FOVs can sometimes result in the majority of filtered correspondences supporting a transformation different from the true one. To address this, the correspondence filtering and transformation estimation steps can be iterated multiple times. In every iteration, the supporting correspondences of the previously calculated transformation are removed from the set of correspondences. This results in a different relative transformation being calculated. In case that the correct result has not been found initially, one additional iteration is usually sufficient. To determine, whether a better alignment has been found in the current iteration, the p2p error metric proposed in \cite{Lehtola2017} for $p=2$ is used. This is defined as
\begin{equation}
    E(p) = \left(\frac{1}{N} \sum_{i=1}^{N} w_i(p) \cdot |u_i - v_i|^p\right)^{1/p}
\end{equation}
with
\begin{equation}
    w_i(p) = \begin{cases}
    1, & \text{if } \|u_i - v_i\|_{p} \leq r_c \\
    0, & \text{else}
    \end{cases},
\end{equation}
where $N$ is the number of points $u_i \in P_s$ of the source point cloud that have a nearest neighbor $v_i \in P_t$ in the target point cloud with a distance below the cut-off radius $r_c$. For each pair of point clouds, the transformation yielding the lowest error metric value is taken. The value of $r_c$ is chosen around the maximal downsampling voxel size.

\subsection{Continuous Registration}

The LiDARs are installed at specific locations in the environment. Their mounting position and orientation e.g. on a pole can be assumed to be kept fixed during a measurement. However, the actual sensor pose is constantly changing due to wind and other effects that result in movements of the pole itself. Since a conventional mast is anchored in the ground and the pole head experiences the greatest translational deflection, we simulate the movement of the pole head by a spherical pendulum with inverted gravity. Thus the result of the initial registration is only valid for the first frame and can not directly be used for the subsequent measurement cycles. However it provides an accurate initial guess for the continuous registration. The point clouds in every frame are again pre-processed and downsampled with a voxel grid filter that approximates the points in every voxel by their centroid.

For the continuous registration a pose graph optimization is performed. Nodes in the graph correspond to sensor poses while edges define the constraints, i.e. the relative transformations, between sensor poses. This optimization results in an estimate of the position and orientation of every sensor that is consistent over all paths between sensors and also between different time steps. The pose graph could contain the data of the whole measurement. In this case it would be constructed from the poses of all sensors at all monitored time steps, resulting in a complete representation but also in a high number of graph nodes. Another downside is that the algorithm could only be used after all data has been recorded in an offline post-processing way resulting in a very high output delay. Thus a series of smaller pose graphs is created from the nodes in a sliding window containing $k_w$ time steps. Because the initial poses are part of the first window (until they leave the window), they are pose graph optimized as well.
\begin{figure}
    \centering
    \begin{tikzpicture}
        \tikzstyle{graph_node} = [draw, circle]
        
        \node [graph_node] (0_0) {};
        \node [graph_node, below of=0_0, yshift=-1cm] (0_1) {};
        \node [graph_node, below of=0_1, yshift=-1cm] (0_2) {};
        \node [below of=0_2] {$t_0$};
        \node [right of=0_0, xshift=0.3cm] (dots_0) {$\ldots$};
        \node [below of=dots_0, yshift=-1cm] (dots_1) {$\ldots$};
        \node [below of=dots_1, yshift=-1cm] (dots_2) {$\ldots$};
        \node [graph_node, right of=dots_0, xshift=0.3cm] (1_0) {};
        \node [graph_node, below of=1_0, yshift=-1cm] (1_1) {};
        \node [graph_node, below of=1_1, yshift=-1cm] (1_2) {};
        \node [below of=1_2] {$t_{n-2}$};
        \node [graph_node, right of=1_0, xshift=0.3cm] (2_0) {};
        \node [graph_node, below of=2_0, yshift=-1cm] (2_1) {};
        \node [graph_node, below of=2_1, yshift=-1cm] (2_2) {};
        \node [below of=2_2] {$t_{n-1}$};
        \node [graph_node, right of=2_0, xshift=0.3cm] (3_0) {};
        \node [graph_node, below of=3_0, yshift=-1cm] (3_1) {};
        \node [graph_node, below of=3_1, yshift=-1cm] (3_2) {};
        \node [below of=3_2] {$t_n$};
        \node [left of=0_0] {$S_0$};
        \node [left of=0_1] {$S_1$};
        \node [left of=0_2] {$S_2$};
    
        \draw[->, dashed] (2_0) -- (1_0);
        \draw[->, dashed] (2_1) -- (1_1);
        \draw[->, dashed] (2_2) -- (1_2);
        \draw[->] (3_0) -- node[text=gray, midway, below]{$T_{0,n}^{0,n-1}$} (2_0);
        \draw[->] (3_1) -- node[text=gray, midway, below]{$T_{1,n}^{1,n-1}$} (2_1);
        \draw[->] (3_2) -- node[text=gray, midway, below]{$T_{2,n}^{2,n-1}$} (2_2);
        \draw[->] (3_0) to [bend right] node[text=gray, pos=0.7, above]{$T_{0,n}^{0,0}$} (0_0);
        \draw[->] (3_1) to [bend right] node[text=gray, pos=0.7, above]{$T_{1,n}^{1,0}$} (0_1);
        \draw[->] (3_2) to [bend right] node[text=gray, pos=0.7, above]{$T_{2,n}^{2,0}$} (0_2);
        \draw[->, dashed] (2_0) to [bend right] (0_0);
        \draw[->, dashed] (2_1) to [bend right] (0_1);
        \draw[->, dashed] (2_2) to [bend right] (0_2);
        \draw[->, dashed] (1_0) to [bend right] (0_0);
        \draw[->, dashed] (1_1) to [bend right] (0_1);
        \draw[->, dashed] (1_2) to [bend right] (0_2);
        \draw[->, dashed] (1_1) to (1_0);
        \draw[->, dashed] (1_2) to (1_1);
        \draw[->, dashed] (1_2) to [bend right] (1_0);
        \draw[->] (3_1) to node[text=gray, midway, left]{$T_{1,n}^{0,n}$} (3_0);
        \draw[->] (3_2) to node[text=gray, midway, left]{$T_{2,n}^{1,n}$} (3_1);
        \draw[->] (3_2) to [bend right] node[text=gray, midway, right]{$T_{2,n}^{0,n}$} (3_0);
        \node [draw, dotted, fit=(1_0)(1_1)(1_2)(2_0)(2_1)(2_2)(3_0)(3_1)(3_2)] {};
        %
    \end{tikzpicture}
    \caption{Pose graph for continuous registration with one sensor stream per row and one time step per column for a sliding window of size $k_w=3$ with $n_L=3$. Solid lines indicate transformations that have to be computed for the current frame. The sliding window is shown as a dotted rectangle.}
    \label{fig:continuous_pose_graph}
\end{figure}
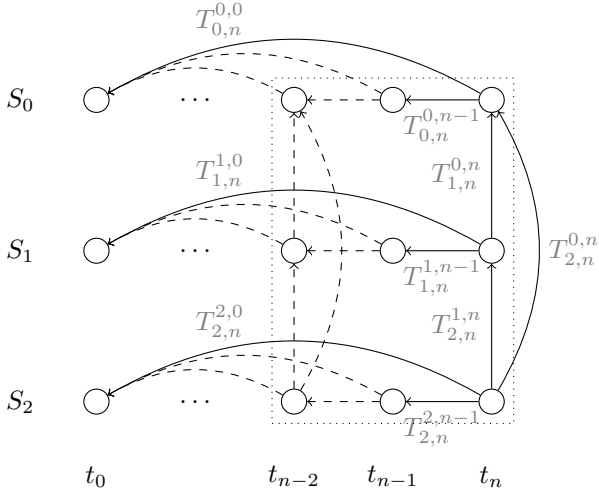
In Fig. \ref{fig:continuous_pose_graph} an example with three sensors is shown. The relative transformations computed at the current time step can be grouped into the following categories:
\begin{enumerate}
    \item Registering against the previous frame for each sensor. This is needed to adapt to sensor movements since the last frame.
    \item Registering against the first frame for each sensor. The additional connections to time step $t_0$ are added to reduce drift over time in the result. Due to dynamic objects in the scene the first and current frame can differ considerably. We deal with dynamic objects in section \ref{subsec:dynamic_objects}.
    \item Pairwise registering of clouds in the current frame between sensors. These edges prevent deviations between the sensor streams from accumulating over time.
\end{enumerate}
The above transformations are determined by employing the G-ICP algorithm that is initialized with already computed transformations from previous time steps. Tab. \ref{tab:continuous_pose_graph_constraints} gives an overview over the initial values used. In case of incomplete data, e.g. due to connection problems, the corresponding fallback values are applied. Edges of type 3) are only calculated and added to the graph every $k_w-1$ steps to reduce runtime. So the initial value $T_{i,n-1}^{j,n-1}$ is usually not available, but can easily be computed from the associated sensor poses.
\begin{figure}
    \centering
    \begin{subfigure}{0.48\columnwidth}
        \includegraphics[width=\columnwidth]{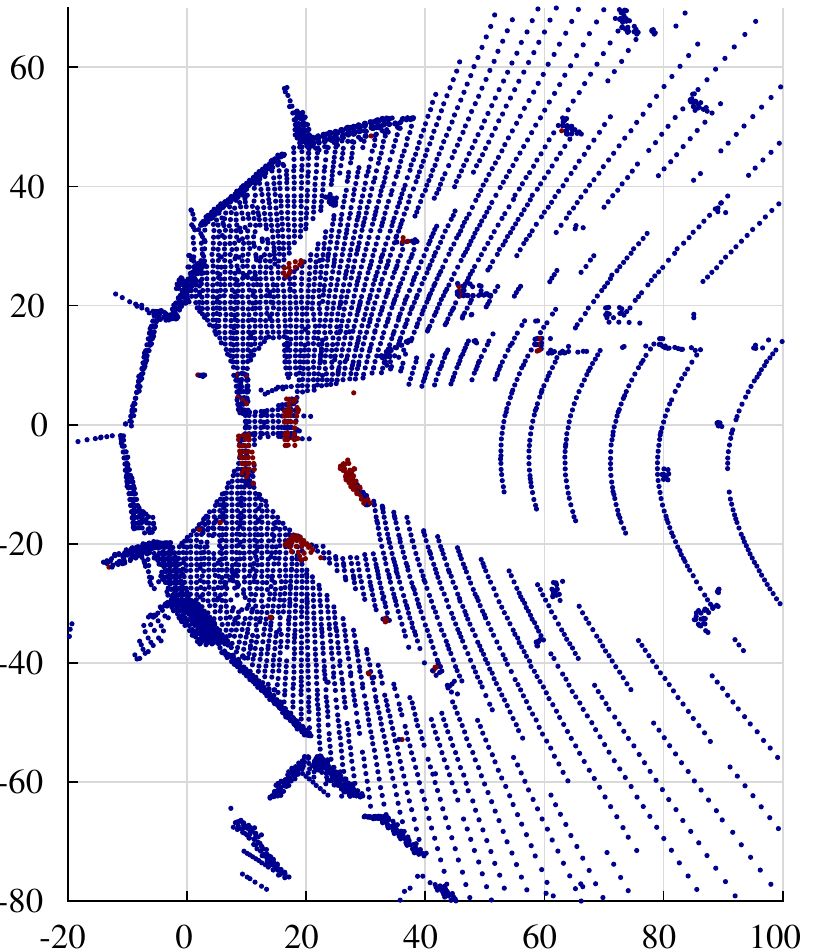}
        \caption{change detections}
    \end{subfigure}
    \begin{subfigure}{0.48\columnwidth}
        \includegraphics[width=\columnwidth]{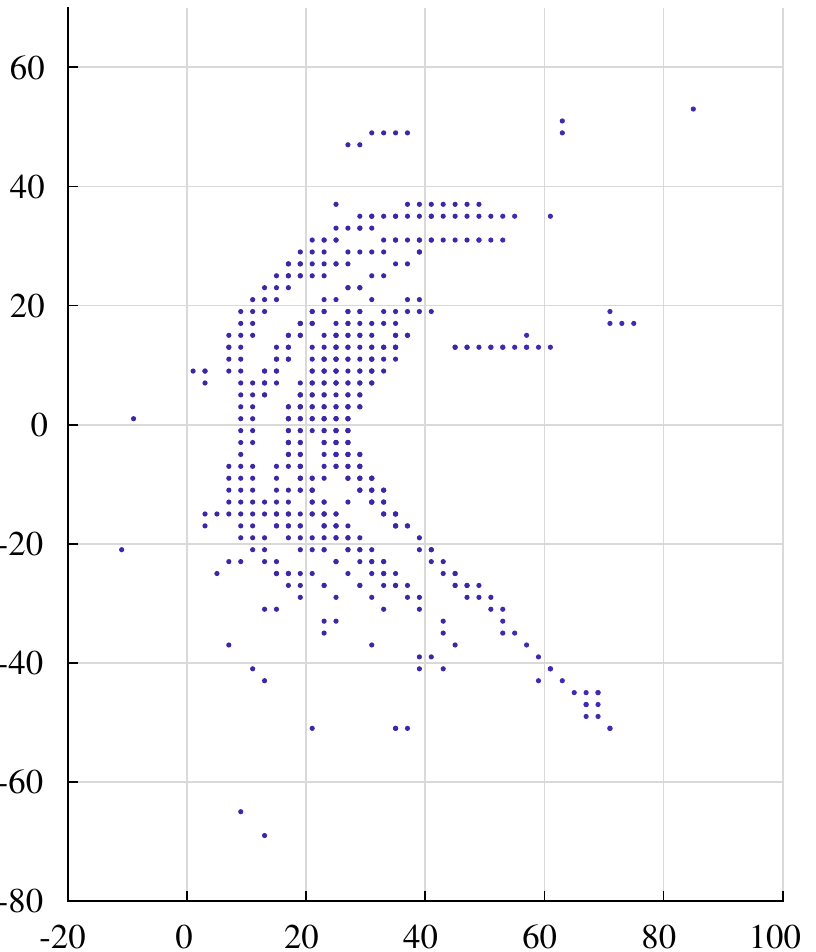}
        \caption{dynamic parts of model}
    \end{subfigure}
    \caption{Segmentation into static and dynamic point sets. Red points in (a) show change detections in an example point cloud. In (b), the resulting model of the dynamic parts of the scene is shown. This corresponds to the street lanes.}
    \label{fig:segmentation_background_model}
\end{figure}
\begin{table}[t]
\caption{Constraints in pose graph.}
\label{tab:continuous_pose_graph_constraints}
\begin{center}
\begin{tabular}{|c|c|c|c|}
\hline
\textbf{Registration target} & \textbf{Transformation} & \textbf{Initial value} & \textbf{Fallback}\\
\hline
Previous frame & $T_{i,n}^{i,n-1}$ & $T_{i,n-1}^{i,n-2}$ & Identity\\
\hline
First frame & $T_{i,n}^{i,0}$ & Identity & Identity\\
\hline
Current frame & $T_{i,n}^{j,n}$ & $T_{i,n-1}^{j,n-1}$ & $T_{i,0}^{j,0}$\\
\hline
\end{tabular}
\end{center}
\end{table}
We employ the Huber loss function to robustify the optimization.
%
%

\subsection{Removal of Dynamic Objects}
\label{subsec:dynamic_objects}

Dynamic objects in the scene like vehicles, pedestrians and other road users can have an impact on the registration result. Also the background is assumed to be static and previously motionless background objects can start moving. As a result, frames of two distinct time steps can differ considerably, impairing the registration. Therefore the continuous registration is applied on the subset of static points only. This is derived from background models. Each sensor holds a separate model that is updated with the dynamic point set of the current time step. The models are attenuated with every frame to reduce the impact of noise and keep the areas of the scene where dynamic points are detected consistently. An example of detected changes and the resulting model are shown in Fig. \ref{fig:segmentation_background_model}.

We employ the change detection algorithm by Underwood et al. \cite{Underwood2013} that works in spherical coordinates and analyzes the neighboring points within a cone around the query point for a significant change in ray length. This way, additions as well as subtractions can be found between two point clouds recorded at different time steps $t_{c_1}$ and $t_{c_2}$ with $c_1 < c_2$. We choose $c_1 = 0$ and $c_2 = n$, so the dynamic points are detected between the first and the current frame at every time step. If we used $c_1 = n-1$ and $c_2 = n$, the model would focus on recently detected changes and lose information of earlier time steps because of the attenuation. The algorithm by Underwood et al. requires already registered point clouds. Because the segmentation is applied before registering the point clouds, the initial value $T_{i,n-1}^{i,0}$ is used to pre-align the scans.

\section{EXPERIMENTS}

\subsection{Simulation Measurements}

\begin{figure}[tb]
    \centering
    \begin{subfigure}{0.32\columnwidth}
        \includegraphics[width=\columnwidth]{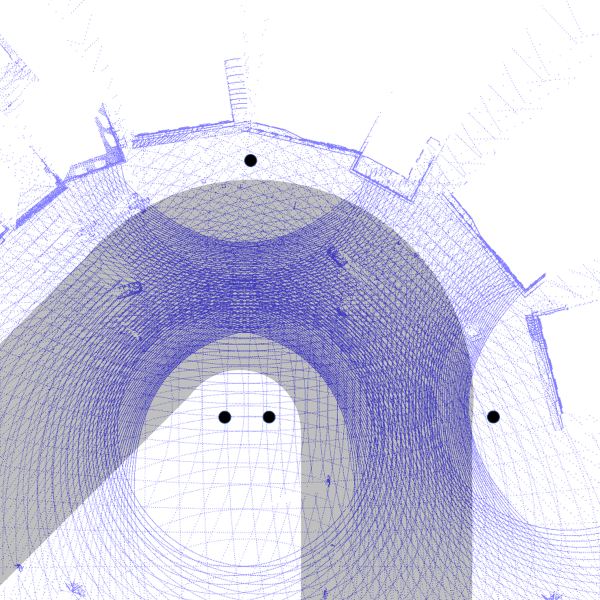}
        \caption{curve}
    \end{subfigure}
    \begin{subfigure}{0.32\columnwidth}
        \includegraphics[width=\columnwidth]{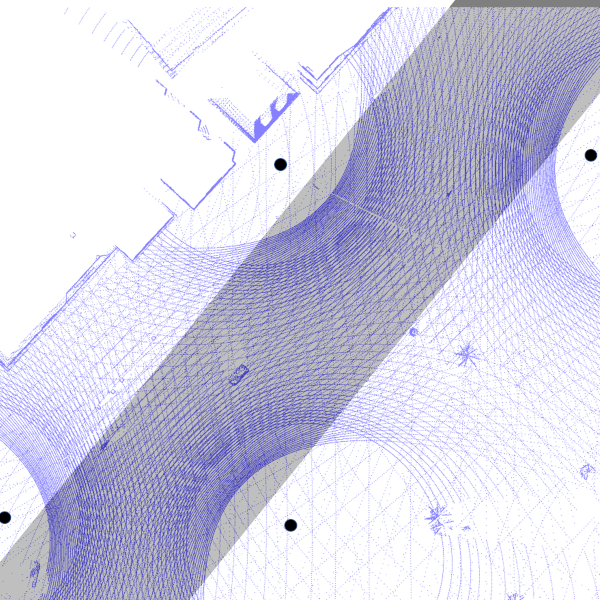}
        \caption{straight}
    \end{subfigure}
    \begin{subfigure}{0.32\columnwidth}
        \includegraphics[width=\columnwidth]{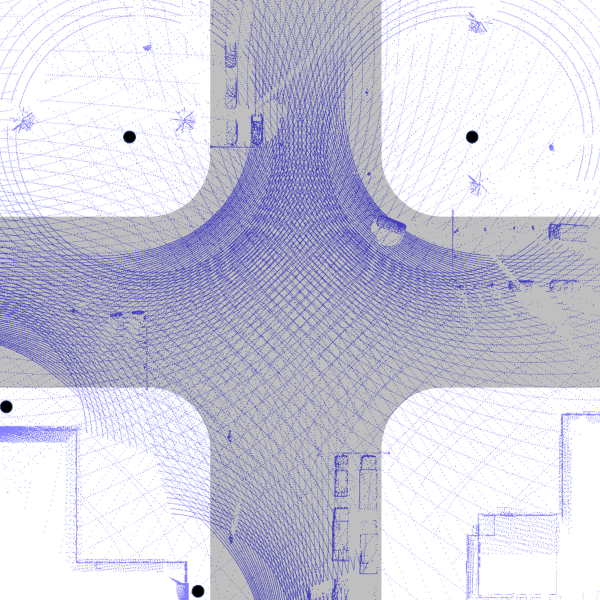}
        \caption{intersection}
    \end{subfigure}
    \caption{Simulated scenarios.}
    \label{fig:vtd_scenarios}
\end{figure}

Different sensor setups (see Fig. \ref{fig:vtd_scenarios}) have been modeled in \emph{Virtual Test Drive (VTD)} \cite{VonNeumann-Cosel2009} using four simulated LiDARs with 64 layers each to evaluate the algorithm's robustness to varying sensor placement. The sensors are in general positioned at a height of six~meters to reduce occlusion and tilted towards the bottom by around \num{17}~degrees. The scenarios contain six traffic lanes (three for each direction). The three setups in detail:
\begin{itemize}
    \item Curve: The LiDARs are positioned along a curve with two sensors on each side of the street. Their viewing direction is perpendicular to the traffic lanes. Buildings and sparse vegetation are present in the scene.
    \item Straight: A straight segment of the road with LiDARs positioned in a zigzag pattern. This results in two sensors being placed on each side of the road respectively. The scene contains buildings and some vegetation.
    \item Intersection: The scene consists of four road segments joining in an intersection. Buildings around the intersection and traffic lights are part of the scene. The direct sight ray between two sensors is obstructed by buildings, which makes the scene more challenging.
\end{itemize}
Although all setups are performed on planar roads it can be assumed that equal recording conditions can be created even on steep roads by adapting the orientation of the sensor system. Simulated road users include cars, trucks, motorcycles, bicycles and pedestrians. For every scenario, noise is added to the measurement. The noise is normal distributed with $\mu = 0$ and $3\,\sigma = 10$~centimeters, which corresponds to the sensor uncertainty.

\begin{figure}
    \centering
    \includegraphics[width=\columnwidth]{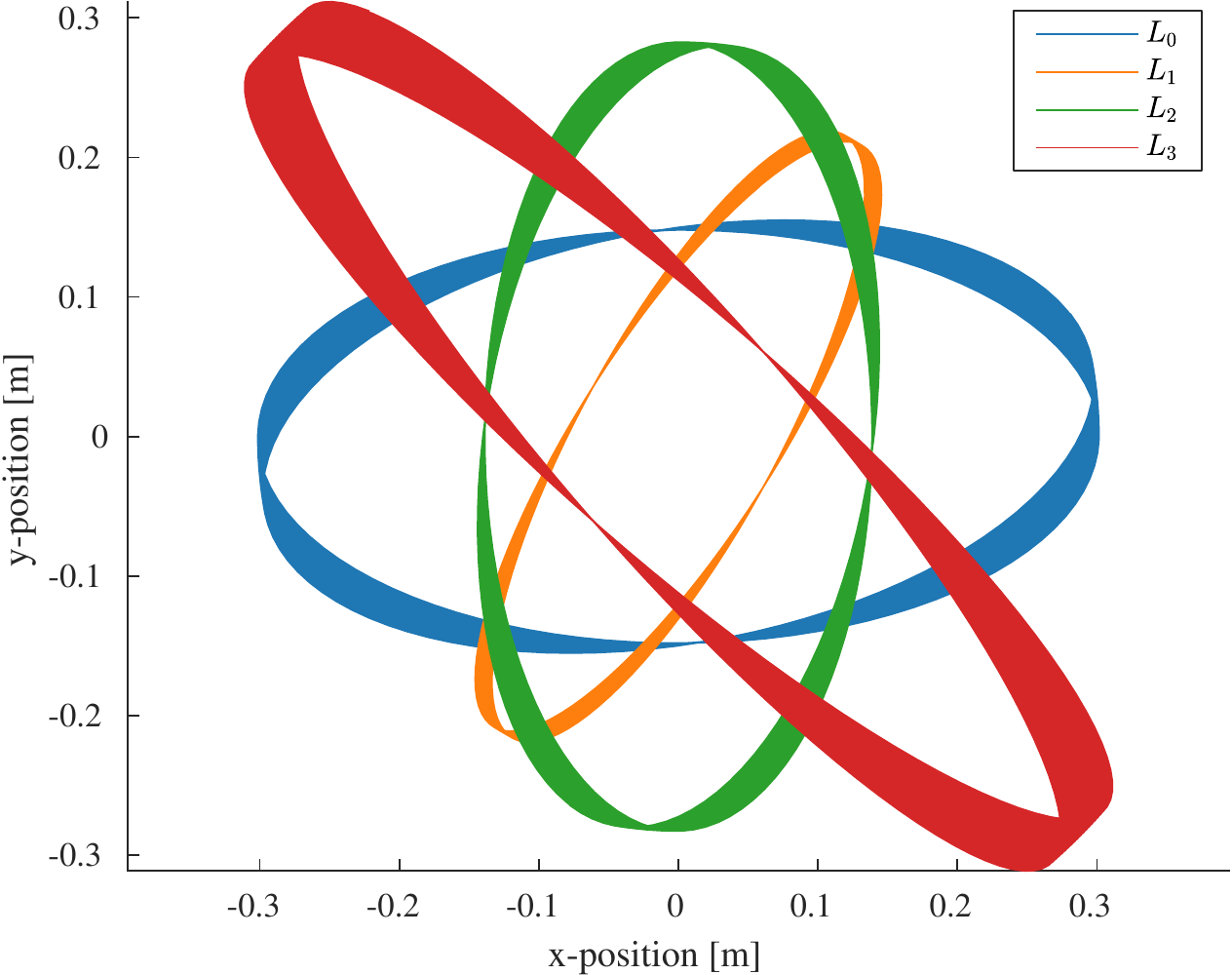}
    \caption{Solution of the spherical pendulum for different sensors plotted for the first \num{300}~seconds.}
    \label{fig:spherical_pendulum}
\end{figure}

The sensor movements are modeled with a spherical pendulum that can be described as
\begin{align}
    \ddot{\theta} &= \dot{\phi}^2 \sin\theta \cos\theta - \frac{g}{r} \sin\theta\\
    \ddot{\phi} &= -2 \, \dot{\phi} \, \dot{\theta} \, \frac{\cos\theta}{\sin\theta}
\end{align}
with polar angle $\theta$, azimuth $\phi$, gravitational acceleration g and radius r (six~meters in our case). The pendulum is initialized with different initial values for each sensor (see Tab. \ref{tab:spherical_pendulum_initial_values}). To ensure comparability of the results, the same initial values are used for the scenarios. Solving the ordinary differential equations at the time steps corresponding to the LiDAR frames yields position and orientation offsets that are added to each sensor pose. The given initial values yield offsets of up to \num{30}~centimeters in x- and y-direction, as shown in Fig. \ref{fig:spherical_pendulum}, which correspond to angular changes of up to four~degrees. In the simulated environment the GT poses as well as the corresponding relative transformations are known for every frame. This allows for an accurate evaluation of the proposed method. With the presented settings of the simulation, which comprise sensor vibrations, sensor noise, feature-rich environments and multiple classes of road users, we expect a complex environment close to the real world for testing and evaluating our algorithm.
\begin{table}
\caption{Initial values for spherical pendulum.}
\label{tab:spherical_pendulum_initial_values}
\begin{center}
\begin{tabular}{|c|c|c|}
\hline
\textbf{Variable} & \textbf{Minimum value} & \textbf{Maximum value}\\
\hline
$\phi$ & $0$ & $0.7 \pi$ \\
\hline
$\partial\phi/\partial t$ & $-0.2 \pi$ & $0.2 \pi$\\
\hline
$\theta$ & $-0.02 \pi$ & $0.02 \pi$ \\
\hline
$\partial\theta/\partial t$ & $-0.01 \pi$ & $0.01 \pi$\\
\hline
\end{tabular}
\end{center}
\end{table}

\subsection{Real-world Measurements}

\begin{figure}
    \centering
    \begin{tikzpicture}
        \tikzstyle{sensor} = [regular polygon, regular polygon sides=3, minimum size=0.3cm, inner sep=0pt]
        \node [anchor=south west, inner sep=0] (image) at (0,0) {\includegraphics[width=\columnwidth]{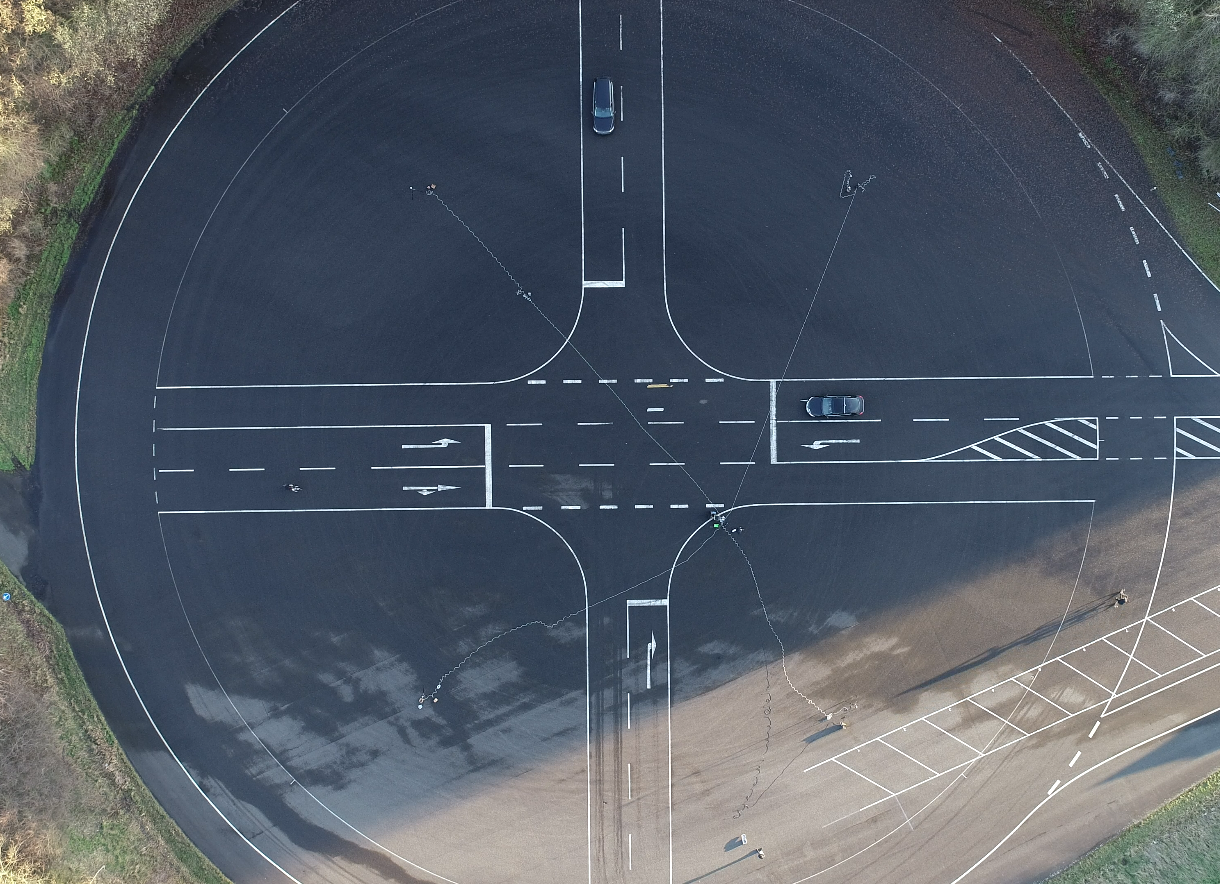}};
        \begin{scope}[x={(image.south east)},y={(image.north west)}]
            \clip (0,0) rectangle (1,1);
        
            
            \draw [fill=green, opacity=0.2] (0.13,0.42) circle (2.29cm); 
            \draw [fill=green] (0.13,0.42) circle (0.05cm);
            \draw [fill=yellow, opacity=0.2] (0.4,0.43) circle (1.48cm); 
            \draw [fill=yellow] (0.4,0.43) circle (0.05cm);
            \draw [fill=white, opacity=0.2] (0.517,0.1) circle (1.61cm); 
            \draw [fill=white] (0.517,0.1) circle (0.05cm);
            \node [draw, sensor, fill=cyan] at (0.34,0.78) {}; 
            \node [draw, sensor, fill=cyan] at (0.695,0.78) {}; 
            \node [draw, sensor, fill=cyan] at (0.345,0.205) {}; 
            \node [draw, sensor, fill=cyan] at (0.7,0.205) {}; 
            \node [draw, sensor, fill=orange] at (0.634,0.568) {}; 
            \node [draw, sensor, fill=orange] at (0.634,0.475) {}; 
            \node [draw, sensor, fill=orange] at (0.7,0.475) {}; 
        \end{scope}
    \end{tikzpicture}
    \caption{Real-world LiDAR setups with road users at our institute's test track in Aachen. LiDAR sensors of the large and small experiment are marked as cyan and orange triangles respectively. Reference points and distances to one sensor of the large experiment are shown as circles. The radius of the inner lane circle is \num{40}~meters.}
    \label{fig:ika_test_track_lidar_setup_far}
\end{figure}
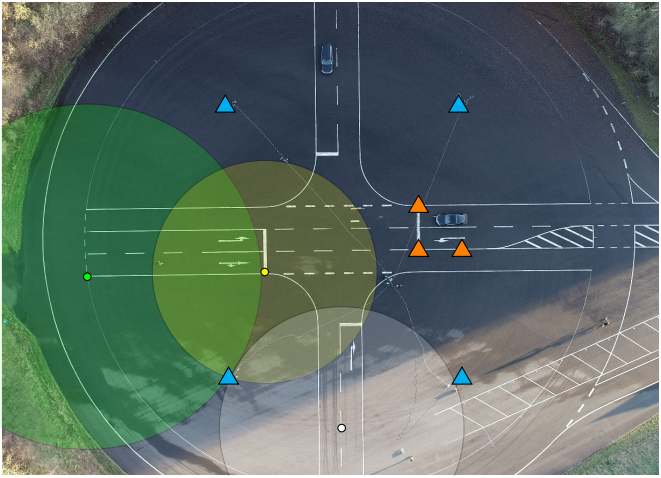

The real-world measurements have been carried out on our institute's test track in Aachen. The test track consists of a circular area with a radius of \num{40}~meters and lane markings of an intersection, surrounded by trees. The experiments in detail:
\begin{enumerate}
    \item Large: Wide sensor setup with four LiDARs, marked as cyan triangles in Fig. \ref{fig:ika_test_track_lidar_setup_far}. The sensors are mounted two~meters high and oriented parallel to the ground plane. Cars, bicycles and pedestrians are in the scene.
    \item Small: Sensor setup consisting of three LiDARs with different inclination angles at a height of two~meters. The sensors are visualized by orange triangles in Fig.~\ref{fig:ika_test_track_lidar_setup_far}. Pedestrians and static objects are in the scene. The sensors are placed at lane markings.
\end{enumerate}
The installation of the sensors at a height of two meters is conditioned by the fact that conventional tripods were used. Unfortunately, there were no lamp posts or traffic light poles on the test track for mounting the sensors at the time of the recordings.

Obtaining a precise GT for the sensor poses in real-world measurements is a challenging task. Since the exact orientation of the sensors was not available, the real sensor measurements are evaluated using the GT data for the sensor position only. The accuracy in orientation estimation and the robustness of the algorithm to sensor motions is evaluated using the VTD scenarios with known GT for position and orientation. The GT for the real-world experiments is determined once for every scenario and kept fixed, because the sensors were not moved during a measurement. For the large-scale experiment, the GT in UTM-WGS84 coordinates was determined using image data from a drone in combination with computer vision algorithms and an orthophoto of the scene. In addition, it was determined as the intersection of three circles derived from the distance to three reference points for every sensor to verify the results (see Fig. \ref{fig:ika_test_track_lidar_setup_far}). Both measurements were averaged to generate the final GT position. For the small-scale experiment the GT in UTM-WGS84 coordinates was generated from the orthophoto as well.

\subsection{Evaluation Metrics}

Each sensor pose in a setup consists of a position and an orientation. The GT position of LiDAR $L_i$ at the current time step $t_n$ is given as $p_{i,n} \in \mathbb{R}^3$. Its orientation is described by a unit quaternion $q_{i,n}$, that is transformed into a matrix $R_{i,n} \in SO(3)$. The predicted sensor position and orientation are denoted by $\hat{p}_{i,n}$ and $\hat{R}_{i,n}$ respectively. However, they are given w.r.t the local sensor coordinate system of $L_0$, whereas the GT is given w.r.t. the world coordinate system. The pose information in the local sensor coordinate system is thus transformed to the world coordinate system for comparisons. The transformed predicted pose is computed as
\begin{align}
    \tilde{p}_{i,n} &= T_{coord} \, \hat{p}_{i,n}\\
    \tilde{R}_{i,n} &= \hat{R}_{i,n} (R_{coord})^{-1},
\end{align}
where the matrix
\begin{equation}
    T_{coord} =
    \begin{bmatrix}
    R_{coord} & t_{coord} \\
    0 & 1
    \end{bmatrix}
    \in SE(3)
\end{equation}
with $R_{coord} \in SO(3)$ and $t_{coord} \in \mathbb{R}^3$ transforms from the local sensor to the world coordinate system. This matrix is unknown. We approximate it from the known point correspondences between sensor positions in both coordinate systems using SVD to find the best alignment in the least squares sense.

The accuracy of the proposed method is measured separately for each sensor w.r.t. the error in translation and rotation given by
\begin{equation}
    E_{trans}(i,n) = \left\|\tilde{p}_{i,n} - p_{i,n}\right\|_2
\end{equation}
and
\begin{equation}
    E_{rot}(i,n) = |\theta_{i,n}| = \arccos\left(\frac{\tr(\tilde{R}_{i,n} R_{i,n}^{-1})-1}{2}\right),
\end{equation}
where $\tr(\cdot)$ is the trace of a matrix and $\theta_{i,n}$ is the rotation angle of the rotation between $\tilde{R}_{i,n}$ and $R_{i,n}$. We do not consider the individual components of the translation and rotation separately as this can easily result in an underestimation of the actual error.

For the synthetic datasets the above error metrics are also evaluated for every frame of the continuous registration. They are combined into a single value for every sensor per scenario using the root-mean-square error (RMSE) calculated as
\begin{equation}
    \rmse_{trans}(i) = \sqrt{\frac{\sum_{n=1}^N (E_{trans}(i,n))^2}{N}}
\end{equation}
and
\begin{equation}
    \rmse_{rot}(i) = \sqrt{\frac{\sum_{n=1}^N (E_{rot}(i,n))^2}{N}}
\end{equation}
for all $i$ sensors of a scenario. The RMSE values of the single sensors are then averaged for every scenario and displayed in Tab. \ref{tab:comparison_accuracy_runtime}.

The runtime is evaluated separately for the initial and continuous registration. The average runtime per frame for the continuous registration is calculated.

\section{RESULTS}

The proposed algorithm is evaluated in two different ways. First, the accuracy that can be obtained with moderate runtime requirements is examined. This is suitable for offline processing tasks. For time-critical applications, the algorithm's real-time capability is evaluated w.r.t. the achievable accuracy under these circumstances. Because of the voxel grid downsampling used for the continuous registration, the voxel size (i.e. the side length of a voxel) is a crucial parameter. A computer with an \emph{Intel Core i9-9900K} CPU was used for the evaluations.
\setlength{\tabcolsep}{5pt}
\begin{table}
\caption{Evaluation of accuracy and runtime.}
\label{tab:comparison_accuracy_runtime}
\begin{center}
\begin{tabular}{|l|c|c|c|c|}
\hline
\multirowcell{2}{\textbf{Scenario}\\ \textbf{(voxel size [\si[detect-weight=true]{\m}])}} & \multicolumn{2}{c|}{\textbf{Runtime}} & \multicolumn{2}{c|}{\textbf{Error}}\\
\cline{2-5}
 & \textbf{Initial [\si[detect-weight=true]{\s}]} & \textbf{Cont. [\si[detect-weight=true]{\ms}]} & \textbf{Trans. [\si[detect-weight=true]{\cm}]} & \textbf{Rot. [\si[detect-weight=true]{\degree}]} \\
\hline
Curve (1) & 10.13 & 288.7 & 2.4 & 0.115\\
\hline
Curve (3) & 10.13 & 93.6 & 9.3 & 0.138\\
\hline
Straight (1) & 11.30 & 321.5 & 5.1 & 0.074\\
\hline
Straight (3) & 11.31 & 93.3 & 14.0 & 0.150\\
\hline
Intersection (1) & 9.94 & 335.9 & 4.0 & 0.030\\
\hline
Intersection (3) & 10.15 & 110.4 & 13.9 & 0.078\\
\hline
Real large (1) & 11.27 & 219.9 & 64.6 & ---\\
\hline
Real large (3) & 11.38 & 52.8 & 72.4 & ---\\
\hline
Real small (1) & 6.13 & 100.3 & 16.2 & ---\\
\hline
Real small (3) & 6.14 & 34.7 & 10.8 & ---\\
\hline
\end{tabular}
\end{center}
\end{table}
\setlength{\tabcolsep}{6pt}

\subsection{Accuracy}

\begin{figure}
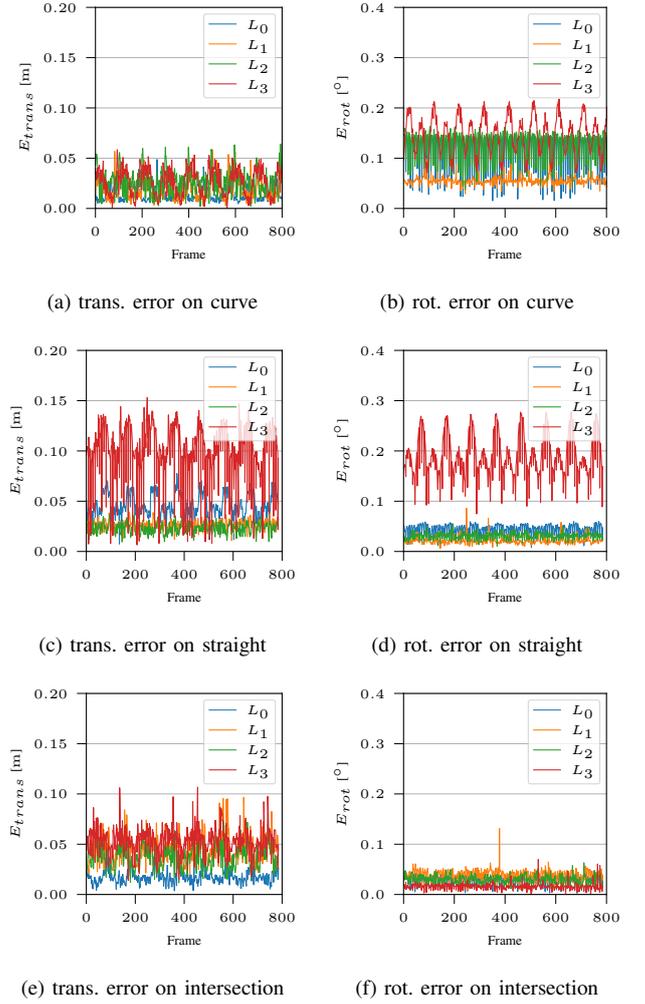

    \begin{center}
        \begin{subfigure}{0.48\columnwidth}
            \input{res/img/identity_1/translational_error_curve_id_1.pgf}
            \caption{trans. error on curve}
        \end{subfigure}
        \begin{subfigure}{0.48\columnwidth}
            \input{res/img/identity_1/rotational_error_curve_id_1.pgf}
            \caption{rot. error on curve}
        \end{subfigure}
    \end{center}
    \begin{center}
        \begin{subfigure}{0.48\columnwidth}
            \input{res/img/identity_1/translational_error_straight_id_1.pgf}
            \caption{trans. error on straight}
        \end{subfigure}
        \begin{subfigure}{0.48\columnwidth}
            \input{res/img/identity_1/rotational_error_straight_id_1.pgf}
            \caption{rot. error on straight}
        \end{subfigure}
    \end{center}
    \begin{center}
        \begin{subfigure}{0.48\columnwidth}
            \input{res/img/identity_1/translational_error_junction_id_1.pgf}
            \caption{trans. error on intersection}
        \end{subfigure}
        \begin{subfigure}{0.48\columnwidth}
            \input{res/img/identity_1/rotational_error_junction_id_1.pgf}
            \caption{rot. error on intersection}
        \end{subfigure}
    \end{center}
    \caption{Error in translation (a, c, e) and rotation (b, d, f) on synthetic datasets for a voxel size of one~meter.}
    \label{fig:error_exact}
\end{figure}
\begin{figure}[h!]
    \centering
    \includegraphics[width=\columnwidth]{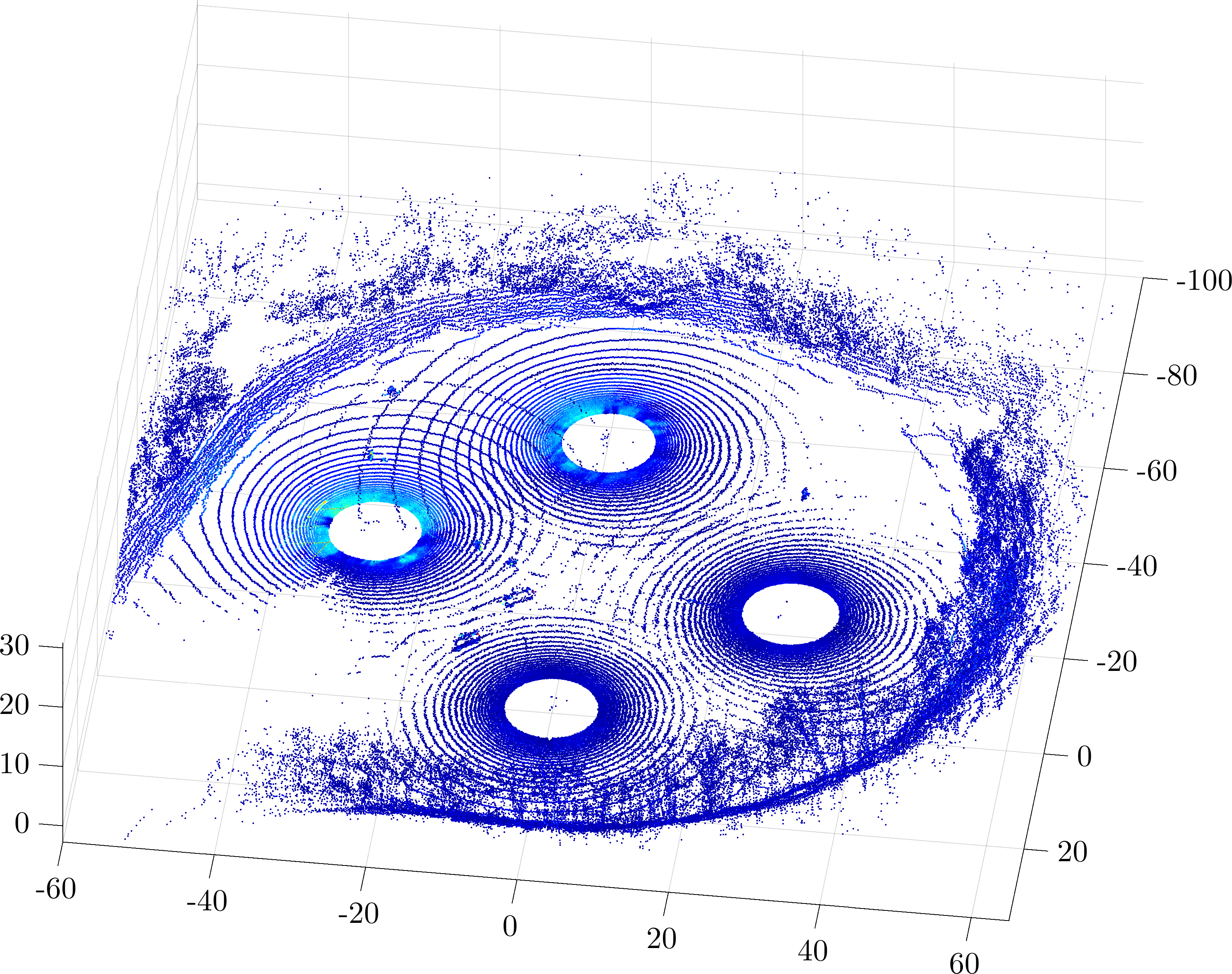}
    \caption{Fusion results on large real-world experiment.}
    \label{fig:fusion_results_real_world_large}
\end{figure}
\begin{figure}[h!]
    \centering
    \includegraphics[width=\columnwidth]{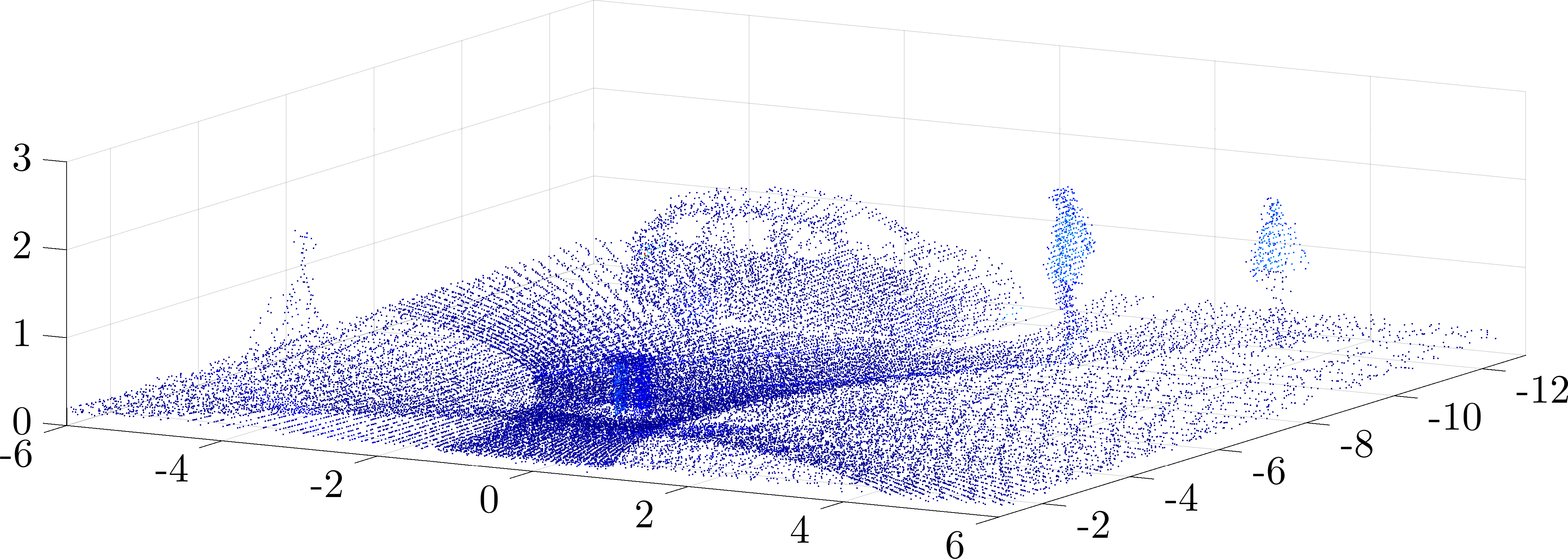}
    \caption{Fusion results on small real-world experiment.}
    \label{fig:fusion_results_real_world_small}
\end{figure}
For the evaluation w.r.t. the achievable accuracy a voxel size of one~meter was used. The errors in translation and rotation of the registration for the simulated datasets are shown in Fig. \ref{fig:error_exact}. We obtain a maximum translational error that is below the artificial sensor noise of ten~centimeters in most cases, except for sensor $L_3$ of the straight scenario. This also yields the maximum error in rotation of \num{0.28}~degrees for LiDAR $L_3$, which is the sensor with the maximum amplitude of the movement (see Fig. \ref{fig:spherical_pendulum}). The maximum averaged RMSE errors of all sensors of a scenario are \num{5.1}~centimeters in translation and \num{0.115}~degrees in rotation for the simulated datasets (see Tab. \ref{tab:comparison_accuracy_runtime}). This is far below the simulated sensor movements (maximum of \num{30}~centimeters in x- and y-direction and up to four~degrees angular change), which are effectively filtered out. Plus, drift is compensated for by the pose graph design and not present in the datasets. 

The evaluation of the challenging real-world datasets shows an increased translational error. Despite this, a qualitative analysis of the resulting point clouds shows a good visual quality as shown in Fig. \ref{fig:fusion_results_real_world_large} and \ref{fig:fusion_results_real_world_small}. Thus, they can still serve as an input to object detection and tracking algorithms. The deviations could at least partially be explained by inaccuracies in the real-world GT, that is hard to estimate exactly. The large real-world experiment is registered with a minimum pairwise sensor distance given. The maximum average runtime per frame with the given voxel size is \num{335.9}~milliseconds.

\subsection{Real-time capability}

\addtolength{\textheight}{-4.3cm}   
%
\begin{figure}
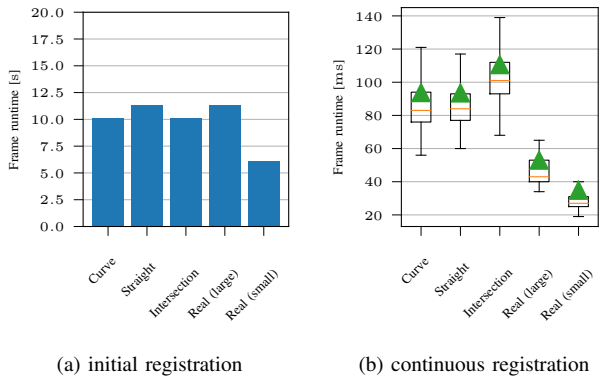

    \centering
    \begin{subfigure}{0.48\columnwidth}
        \input{res/img/identity_3/runtime_initial.pgf}
        \caption{initial registration}
    \end{subfigure}
    \begin{subfigure}{0.48\columnwidth}
        \input{res/img/identity_3/runtime_continuous.pgf}
        \caption{continuous registration}
    \end{subfigure}
    \caption{Runtime of the initial and continuous registration for a voxel size of three meters. The average runtime per frame is marked with a green triangle in (b).}
    \label{fig:runtime_fast}
\end{figure}
Increasing the voxel size allows for a faster processing of the downsampled point clouds that qualifies the algorithm to run in real-time. With a voxel size of three~meters the average runtime is below \num{100}~milliseconds for most simulated scenarios and below \num{53}~milliseconds on real-world data (see Fig. \ref{fig:runtime_fast}). The algorithm is thus capable to process data at a sampling rate of ten~hertz in real-time. Because most LiDARs support a sampling rate between \num{5} and \num{20} hertz, this is sufficient for real-time applications.
As a trade-off, the registration errors increase to a maximum error of \num{14}~centimeters and \num{0.15}~degrees in translation and rotation for the simulated datasets. The alignments still provide a high-quality input to subsequent analysis. This can be assessed qualitatively in Fig. \ref{fig:comparison_fusion_result_and_gt}, where the fusion result with a voxel size of three~meters and the GT are shown next to each other for one simulated scenario.

%
%

\section{CONCLUSIONS}

Our presented algorithm allows a point cloud registration of up to four 64-layer LiDARs by automatically determining the extrinsic calibration parameters. It has successfully been applied to simulation and real-world data in real-time. Our experiments have shown that the average resulting registration errors in translation and rotation are sufficiently low to have no apparent effect on the fused point clouds. The developed approach therefore provides an optimal sensor data basis for the use of deep learning algorithms for high-precision detection of road users. In future work we want to scale our approach in terms of sensor number and sensor quality and prove it in large-scale test fields.








\bibliographystyle{IEEEtran} 
\bibliography{literature}

\end{document}